\documentclass[sn-mathphys,Numbered, iicol]{sn-jnl}

\usepackage{graphicx}%
\usepackage{multirow}%
\usepackage{amsmath,amssymb,amsfonts}%
\usepackage{amsthm}%
\usepackage{mathrsfs}%
\usepackage[title]{appendix}%
\usepackage{xcolor}%
\usepackage{textcomp}%
\usepackage{manyfoot}%
\usepackage{booktabs}%
\usepackage{algorithm}%
\usepackage{algorithmicx}%
\usepackage{algpseudocode}%
\usepackage{listings}%
\usepackage{tikz}
\usepackage{pgf-pie}
\usepackage{xcolor,colortbl}
\usepackage{flushend}

\theoremstyle{thmstyleone}%
\newtheorem{theorem}{Theorem}
\newtheorem{proposition}[theorem]{Proposition}%

\theoremstyle{thmstyletwo}%

\theoremstyle{thmstylethree}%
\begin{document}

\title[Article Title]{Explaining Autonomy: Enhancing Human-Robot Interaction through Explanation Generation with Large Language Models}

\author*[1]{\fnm{David} \sur{Sobrín-Hidalgo}}\email{dsobh@unileon.es}

\author[1]{\fnm{Miguel A.} \sur{González-Santamarta}}\email{mgons@unileon.es}

\author[1]{\fnm{Ángel M.} \sur{Guerrero-Higueras}}\email{am.guerrero@unileon.es}

\author [1]{\fnm{Francisco J.} \sur{Rodr{í}guez-Lera}}\email{fjrodl@unileon.es}

\author[1]{\fnm{Vicente} \sur{Matellán-Olivera}}\email{vmato@unileon.es}

\affil[1]{\orgdiv{Robotics Group. Department of Mechanical, Computer and Aerospace Engineering}. \orgname{University of León}, \orgaddress{\street{Campus de Vegazana s/n}, \city{León}, \postcode{24007}, \state{Castilla y León}, \country{Spain}}}




\abstract{This paper introduces a system designed to generate explanations for the actions performed by an autonomous robot in Human-Robot Interaction (HRI). Explainability in robotics, encapsulated within the concept of an eXplainable Autonomous Robot (XAR), is a growing research area. The work described in this paper aims to take advantage of the capabilities of Large Language Models (LLMs) in performing natural language processing tasks. This study focuses on the possibility of generating explanations using such models in combination with a Retrieval Augmented Generation (RAG) method to interpret data gathered from the logs of autonomous systems. In addition, this work also presents a formalization of the proposed explanation system. It has been evaluated through a navigation test from the European Robotics League (ERL), a Europe-wide social robotics competition. Regarding the obtained results, a validation questionnaire has been conducted to measure the quality of the explanations from the perspective of technical users. The results obtained during the experiment highlight the potential utility of LLMs in achieving explanatory capabilities in robots.}




\maketitle

\section{Introduction}\label{sec:introduction}

The discourse on Human-Robot Interaction (HRI) is inseparable from the concept of trustworthiness; thus, researchers in social robotics are investing great efforts in improving human trust in robots \cite{trustworthy2015}. HRI is a field where interactions between humans and robots are constant. For this reason, humans must rely on the information provided by robots, as well as on their decision-making capabilities. However, for humans to trust such information, they must first understand it \cite{trustworthy_explainability}. In this context, the importance of explainability becomes evident. An explainable system should provide human-understandable explanations about the behavior executed, the decisions made, or any events that occurred during the execution.

For example, imagine a simple navigation task where the robot is required to move to a specific room in a house. However, all doors are closed, and there is no available path. A user may be interested in understanding why the robot cannot reach its goal. This type of information is precisely what explainability aims to provide.

The area of robotic explainability is trying to address some issues associated with legal and security matters. It is undeniable that there are ethical implications in social robotics, which arise from the continual interaction between humans and robots. The presence of humans in the robot environment requires the implementation of specific safety measures, essential for enhancing trust. Achieving optimal safety levels demands the establishment of mechanisms that allow systems to trace and record any potential failure or accident. Thus, users can understand the origin and cause of any relevant event. These facts lead to the conclusion that autonomous robots must allow the traceability of their decision-making processes and the events that arise during the execution of their behavior. Simultaneously, the data recorded must be interpretable by human users.

Another important aspect involves the definition of the concepts of explainability and explanation. There is no consensus about these definitions within the literature. In \cite{definiciones2019}, the authors assert that an explainable system is closely related to the concepts of interpretability and transparency. Therefore, an explainable system must also be supported by a robust system of accountability. To generate an explanation, relevant data must be gathered, and that data must be interpretable.

Regarding the definition of explanation, as suggested by the authors in \cite{definiciones2019} based on Van Fraassen's work \cite{van198511}, an explanation is an answer to "Why" questions in which one option is compared to another. This conceptualization has evolved over the years, to the point that an explanation should be able to answer a wider variety of questions (WHY, WHO, HOW, WHAT).

On the other hand, within the same study, and referencing Achinstein's work in \cite{achinstein1983nature}, authors propose that an explanation must involve an exchange of knowledge between two actors. This implies that answering the question is insufficient, but that the response must result in knowledge acquisition for the receiver. This idea underscores the need for the explanation to be understandable by the receiver.

These definitions highlight the importance of the process by which explanations are generated. However, the influence of the concepts of understandability and interpretability underscores the need to consider the user's knowledge level when defining the concept of explanation. Transferring this idea to the field of information technology in general, and robotics specifically, different types of users can be distinguished, such as technical users and end users, for instance.
As a result of the difference between users' levels of knowledge, certain data may be easily interpreted and understood by users who possess a higher level of specialization while proving challenging for those unfamiliar with the technical aspects of the system.

From all these concepts and proposals, and with a specific emphasis on HRI, the following definition will be used to lead the rest of the work:

\begin{proposition}
An explanation is an answer constructed from interpretable data. It must be understandable by the user and improve the user's understanding of the system's behavior.
\end{proposition}

From this definition, the following procedure has been established to generate explanations:

\begin{enumerate}
\item \textbf{Collecting relevant data}: Gathering data concerning the events occurring within the system during its execution.
\item \textbf{Interpretation of data}: Ensuring the data is interpretable.
\item \textbf{Explanation construction}: Using the interpretable data to construct the explanation, considering account the user's knowledge level.
\end{enumerate}

This work proposes to address this procedure through the use of \textit{Large Language Models} (LLMs) to develop an interactive explainability system. The context chosen for this study will focus on the navigation skill of a robot operating in an environment where obstacles may arise, causing path replanning. The LLM assumes a dual role within the explainability system. On the one hand, it serves as an interpreter for the collected data. On the other hand, it functions as an explanation manager, similar to a dialog manager, responsible for generating appropriate responses to user questions regarding the robot's behavior.

The system's interactivity is achieved through two key elements: first, the process of exchanging questions and answers between the user and the robot; and second, the capability to provide real-time responses. The research conducted in this paper aims to check if it is possible to achieve the desired level of interactivity.
Concerning the accountability system, an approach based on the analysis of the robot log files has been proposed. The logs produced by ROS 2 will be collected and enhanced with some custom logs defined by the developers. The primary objective of these additional logs is to increase the verbosity of the results obtained from the ROS 2 log system.

Based on this approach, this work focuses on answering the following research questions:

\begin{description}
\item[\textbf{RQ1}] Can LLMs be used to generate explanations for autonomous robots' behavior by analyzing the logs they generate during their tasks?
\item[\textbf{RQ2}] Are the logs produced by an autonomous robot sufficient to generate an explanation, or should the verbosity of the logs be increased?
\end{description}

This article is structured as follows. First, a comprehensive exploration of related works is presented in Section \ref{sec:relatedWorks}. Section \ref{sec:arch} defines the architecture proposed for the system, and Section \ref{sec:materials} details the implemented software solution along with the experimentation design and the definition of the metrics used. Section \ref{sec:results} showcases the acquired results, which are discussed and analyzed in Section \ref{sec:discussion}. Finally, Section \ref{sec:conclusions} presents the overall conclusions of the work.

\section{Related works}\label{sec:relatedWorks}

The purpose of this section is to summarize previous works related to the explainability system presented in this article. We begin from a theoretical perspective, reviewing works within the domain of robotics explainability.

Explainability is a broad and complex concept. This study focuses on two branches: eXplainable Artificial Intelligence (XAI) and eXplainable Autonomous Robots (XAR). XAI explains the decision-making processes of machine learning models, often perceived as "black boxes" by users. Meanwhile, XAR aims to provide explanations for the behavior of autonomous robots. Thus, XAR can be seen as encompassing, at least partially, the domain of XAI. This is because robot behavior often involves the use of artificial intelligence mechanisms in skills such as object or voice recognition.

The research community has devoted significant efforts to exploring XAI, focusing on developing various algorithms and frameworks to analyze the reasoning behind a model's predictions. Literature reviews in XAI, such as \cite{vilone2020explainable}, focus on gathering different methods proposed for generating explanations, particularly in applications like computer vision and classification problems. A popular technique involves perturbing the model input and observing its effect on the final prediction. These algorithms' outputs are often presented as heat maps or graphs, indicating the influence of different model features on the prediction.

DARPA's ``Explainable Artificial Intelligence Program'' \cite{gunning2019darpa}, dedicated to studying explainability within Machine Learning and Deep Learning systems, has produced a literature review \cite{mueller2019explanation} that consolidates concepts and definitions from a theoretical perspective within XAI.
Conversely, the XAR area has received relatively less attention, possibly due to its broader and less specific goals compared to XAI. The literature review in \cite{review_xar_2019} examines how explanations for robot actions can be presented and evaluated at a theoretical level. It analyzes domains where explainability systems are applied and platforms used for designing these explanations.

A wide variety of works address the generation of explanations for robot behavior. These approaches vary significantly, focusing on explanations for specific robot skills. Given the diverse spectrum of abilities in robot behavior, the mechanisms used to generate explanations can differ considerably. Consequently, this section will now focus on discussing some technical approaches proposed in the literature.

The study depicted in \cite{rosenthal2016verbalization} introduces an algorithm allowing robots to verbalize their navigation routes, adjusting the level of detail. This solution enables robots to narrate their real-world navigation experiences in natural language.

Another relevant approach for this paper views explanations as the result of a dialogue between the system and the human user. Various studies exploring this perspective have been reviewed. In \cite{bansal2018explanatory}, a framework for explanatory dialogues is defined. \cite{arioua2015formalizing} introduces a dialog-based explanation system named ECE. Although these proposals are oriented towards XAI, the concept of explanations through interactive dialogue can be extrapolated to robotics.

Our previous work in \cite{fernandez2023accountability} introduces a system for analyzing data in robot log files, storing them in a black box component, which enhances record security. It also proposes algorithms to produce natural language explanations based on collected data.
These examples highlight the potential of dialogue managers in generating explanations and the importance of data verbosity, which increases interpretability.

In recent years, LLMs have gained considerable attention in both, research and business sectors. As a subset of deep learning models, LLMs have shown significant potential for text generation and solving Natural Language Processing (NLP) tasks. Their widespread adoption in applications like intelligent chats, exemplified by OpenAI's ChatGPT, demonstrates the practical utility of LLMs as effective dialogue managers. Microsoft's project using LLMs like ChatGPT for robotics applications \cite{microsoftchatgpt} highlights their potential in robotics, even though it does not address explainability.

Other works explore using LLMs to generate natural language explanations. In \cite{tursun2023selfexplainability}, the authors propose combining heat maps with ChatGPT. First, they extract context information from the maps and then use ChatGPT to answer questions related to a specific heat map.

Some literature is based on the concept of explainability systems as dialog managers. We previously explored in \cite{gonzalez2023using} the ability of LLMs (GPT3.05, GPT 4.0, and Alpaca) to interpret logs generated by autonomous robots, using the logs directly without any processing or prompt engineering.

The study outlined in this paper integrates the ideas discussed in this section. The main emphasis is on using an LLM as an interpreter for autonomous robot log files and as a manager of an interactive dialogue based on question-answer communication.

\begin{figure*}[h]%
\centering
\includegraphics[width=0.9\textwidth]{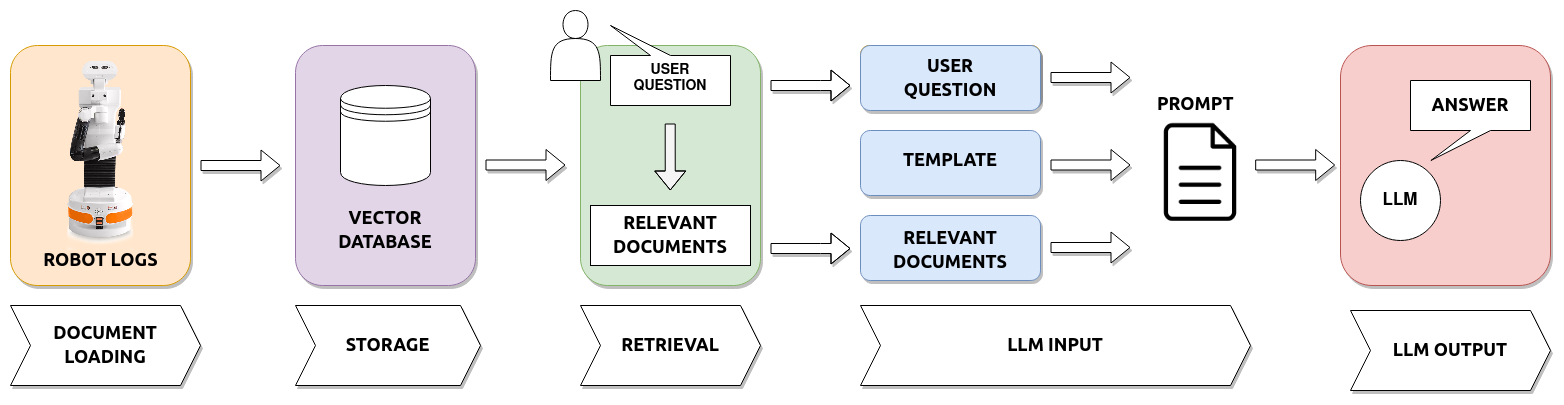}
\caption{Explainability system workflow. Use of robot's logs and RAG method to generate explanations.}\label{fig:rag_diagram}
\end{figure*}

\section{Proposed System Architecture}
\label{sec:arch}

The system we introduce in this paper is designed to interpret logs generated by an autonomous robot and to generate explanations from these logs in response to questions from human users. This section describes the workflow of the proposed system, illustrated in Figure \ref{fig:rag_diagram}, which highlights how a Retrieval Augmented Generation (RAG) method is used within our explainability system.

Robot logs represent various data that can be more or less interpretable depending on the user's level of knowledge. These logs are used to provide a broader context for the LLM, which then interprets these logs to respond to users' queries about this context.

Our system is designed as a software component that operates independently of the robot's specific functions. This means the system can work with any set of logs, regardless of which part of the robot's subsystem produced them. Such a design makes the system flexible, enabling it to handle diverse tasks -- like navigation, manipulation, or object recognition -- without needing significant changes.

Our system employs the RAG method \cite{lewis2021retrievalaugmented}. This approach allows the LLM to utilize the logs as context by leveraging a knowledge base, which we implement as a vector database, to broaden the context accessible to the model. When an input query is received, the method retrieves a set of relevant documents from this database. These documents then form the final prompt for the LLM.

As the robot produces logs, the system converts these logs into embeddings, which are vector representations of the data, and stores them in the vector database. Upon receiving a query, the system identifies the most relevant documents from this database through a retrieval process. The generated response is based on these documents, providing contextual information to the user's query.

\bmhead{RAG method formalization} 

We pose a theoretical model of the proposed system, which primarily comprises two elements: a vector database and an LLM. A significant part of the system's complexity lies in managing the robot's logs through the database. Therefore, we will focus on formalizing the relationships within this database.

The system \(S\) is defined as a tuple of two elements, as shown in Equation \ref{eq:system}.

\begin{equation}
\label{eq:system}
S = \langle L, E \rangle
\end{equation}

Here, \(L\) represents the set of logs generated by the robot; and \(E\) represents the set of embeddings.

The set \(L\) of logs is defined in Equation \ref{eq:conjuntoLog}. Each log \(l_i\) in this set, as shown in Equation \ref{eq:elementoLog}, is characterized by two attributes: a timestamp (t) and a string message (m).

\begin{equation}
\label{eq:conjuntoLog}
l_{i} \in L = \{l_{1}, l_{2},...., l_{n}\}, n \in \mathbb{N}
\end{equation}

\begin{equation}
\label{eq:elementoLog}
l_{i} = \langle t, m \rangle
\end{equation}

Correspondingly, each embedding \(e_{i}\) in the set \(E\) (Equation \ref{eq:conjuntoEmb}) is derived by applying the embedding function \(f_{emb}\), defined in Equation \ref{eq:funcionEmb}, to the respective log \(l_{i}\) in set \(L\).

\begin{equation}
\label{eq:conjuntoEmb}
e_{i} \in E = \{e_{1}, e_{2},...., e_{n}\}, n \in \mathbb{N}
\end{equation}

\begin{equation}
\label{eq:funcionEmb}
f_{emb}: l_{i} \rightarrow e_{i}
\end{equation}

The system uses a similarity query to retrieve the \(k\) most relevant log elements  \(l'_{i}\), forming a new subset \(L'\) as indicated in Equation \ref{eq:conjuntoRelevantes}. The subset \(L'\) is obtained using the retrieval function \(f_{ret}\) (Equations \ref{eq:sigma} and \ref{eq:ret}), which applies the Maximal Marginal Relevance (MMR) method to the initial log set \(L\). This function takes the log set \(L\), a user-entered query \(q\), and the number of relevant elements \(k\) as inputs.

\begin{equation}
\label{eq:conjuntoRelevantes}
l'_{i} \in L' = \{l'_{1}, l'_{2},...., l'_{k}\}, k \in \mathbb{N}
\end{equation}

\begin{equation}
\label{eq:sigma}
    L' = \sigma_{f_{ret}}(L)
\end{equation}

\begin{equation}
\label{eq:ret}
f_{ret}: q \times k \times L \rightarrow L'
\end{equation}

Finally, the LLM prompt \(Q\) is formulated as shown in Equation \ref{eq:query}, where \(tp\) is a predefined template, \(q\) is the user's query, and \(L'\) consists of the relevant logs providing context to the LLM.

\begin{equation}
\label{eq:query}
Q = \langle tp, L', q \rangle
\end{equation}
\newline
This section has outlined the theoretical framework of the system. The next section will delve into the technologies used for implementation, offering a more technical view of the system. For the first approach, the experiment has been centered around a navigation test, as navigation is a fundamental skill in social robotics.

To address the navigation test mentioned earlier, the necessary software has been developed. This software not only manages the robot's behavior required to complete the navigation test but also generates customized logs added by the developer. These logs are designed to reinforce the information provided by the robot's log system. The reason behind this implementation resides in the necessity to supply the LLM with sufficiently verbose information.

\section{Materials and methods}\label{sec:materials}

The present section describes the proposed software solution for the explainability system. Additionally, it includes the design of the experimentation used to test the developed system and also the metrics defined for the experiment.

\subsection{System implementation proposal}

The system we have developed is designed to operate within robotic systems that use ROS 2 (Robot Operating System 2) \cite{macenski2022robot}, the leading middleware standard for robotic programming. Consequently, the source code is written in Python, a language supported by ROS 2, according to  the requirements of this middleware.

\begin{figure*}[h]%
\centering
\includegraphics[width=0.9\textwidth]{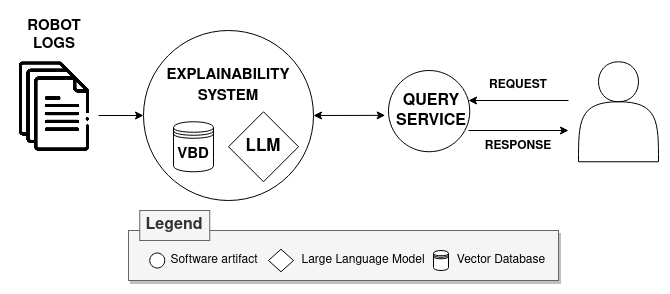}
\caption{Explainability framework components}\label{fig:systemDefinition}
\end{figure*}

As mentioned in the paper's introduction, the primary experiment involves a navigation test where the robot is tasked with moving to three specified positions (waypoints) within an apartment. These waypoints, defined by 7 coordinates (a vector3 for position and a quaternion for rotation), were navigated using Nav2 \cite{macenski2020marathon2}, ROS 2's default navigation framework. Additional software was developed to provide the robot with the necessary navigation capabilities.

To integrate LLM technology into ROS 2 for generating explanations, we employed LLama\_ros \cite{Gonzalez-Santamarta_llama_ros_2023}. This tool facilitates the incorporation of \textit{llama.cpp} into ROS 2 projects.

Several third-party components were also incorporated into our experiment:
\begin{itemize}
    \item \textbf{Vector database}: We selected \textbf{ChromaDB \cite{Chromacore}} for its ability to operate locally.
    \item \textbf{LLM}: The model \textbf{openhermes-2.5-neural-chat-7b-v3-1-7b.Q4\_K\_M.gguf} was chosen for its high ranking on the HuggingFace leaderboard \cite{open-llm-leaderboard} as of December 4, 2023.
    \item \textbf{Langchain \cite{Chase_LangChain_2022}}: This framework is tailored for creating language model-based applications, offering seamless integration with vector databases like Chroma and facilitating the implementation of the RAG method.
\end{itemize}

For the experiment designed to assess the explainability system, a real robot was deployed in an environment resembling real-world conditions. The materials used include:

\begin{itemize}
    \item \textbf{TIAGo}: A mobile manipulator robot from PAL Robotics. 
    \item \textbf{Leon@Home\footnote{More information on the testbed can be found on its official webpage: \url{https://robotica.unileon.es/index.php?title=Testbed}}}: A testbed certified by the European Robotics League (ERL).
    \item \textbf{Laptop}\footnote{The hardware specification is important as the resources impact the model's response time.}: Model Acer Predator Triton 300 SE with Ubuntu 22.04. CPU 11th Gen Intel® Core™ i7-11370H @ 3.30GHz $\times$ 8. RAM 16 GB. GPU Nvidia GeForce RTX 3060 Mobile.
\end{itemize}

\begin{figure*}[h]%
\centering
\includegraphics[width=0.9\textwidth]{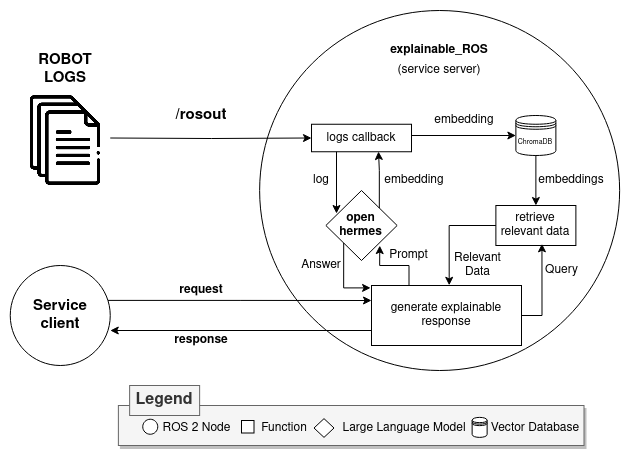}
\caption{Implementation of an explainability system based on the proposed framework.}\label{fig:implementation}
\end{figure*}

Building on the architectural workflow proposed in section \ref{sec:arch} and using the aforementioned technologies, we constructed the explainability system. Figures \ref{fig:systemDefinition} and \ref{fig:implementation} depict the system's high-level definition and technical design, respectively.

Logs generated during the robot's operation are collected to gather data evidencing its behavior. These logs, produced by ROS 2's logging systems and published on the \textit{\/rousout} topic, are complemented by additional messages from the development of the navigation node (\textbf{navigation\_waypoint}) to increase log verbosity and enhance LLM comprehension of events.

For Llama\_ros configuration, several parameters were adjusted: \textbf{n\_threads} for parallel processing management; \textbf{n\_ctx} sets the maximum context size in tokens for the model; \textbf{n\_batch} defines the inference batch size; and finally, \textbf{n\_gpu\_layers} allocates a certain number of layers in Video Random Access Memory (VRAM).

Sampling parameters to response generation include: \textbf{n\_prev}, indicating the number of previous tokens to consider during sampling; \textbf{top\_k}, limiting the selection to the top probable tokens at each step; \textbf{penalty\_last\_n}, to avoid repetitive outputs by penalizing the last n tokens; and \textbf{top\_p} and \textbf{temp} control the randomness of the language model's output. 

In operation, all logs are published on ROS 2's \textit{/rosout} topic. The explainability node, \textbf{explainable\_ROS}, reads and processes these logs, managing user queries through a ROS 2 service. An LLM function calculates embeddings from the data, which are then stored in the vector database. When a user inquires about the robot's behavior, the system retrieves the most relevant logs, creating a context that, along with a predefined template\footnote{Prompt template can be found in listing \ref{lst:template} of Appendix\ref{secA0}} and the user's query, forms the final LLM prompt. The LLM evaluates this query and returns an explanation to the user through a ROS 2 service.

\subsection{Challenges}
\label{sub:desafios}

Throughout the development of our explainability system, we've identified several challenges that mainly affect the performance and quality of the LLM's responses. These challenges are as follows:

\bmhead{Real time}

Our system is designed to calculate embeddings for each log as they are published on the \textit{/rosout} topic. However, generating embeddings requires computation time, which can increase with longer messages. The challenge arises from the disparity between the processing time for each log and the significantly higher publication rate on \textit{/rosout}, creating a bottleneck in the callback function and potentially leading to message loss.

To address this, we implemented a message queue processed sequentially. While this approach does not enable real-time processing, it effectively prevents information loss, marking a satisfactory solution for our system prototype.

\bmhead{Bad retrievals}

The frequent repetition of certain log messages, such as \textit{Passing new path to controller} from Nav2, was found to negatively impact the model's responses. Such repetitiveness does not add relevant information and can degrade the retrieval process's effectiveness. Our solution filters out consecutive messages with identical content, ensuring a message is processed only if it differs from the preceding one. This strategy has improved processing efficiency and the quality of the model's responses.

\bmhead{Retrieval order}

The challenge here is that the most relevant messages retrieved are not presented chronologically, complicating answers to questions about the sequence of events. To provide temporal context, timestamps were included with log messages. Additionally, we applied a simple Python function to sort the context messages by their timestamps, ensuring the LLM considers events in their correct chronological order.

\bmhead{Log's verbosity and question's quality}

The detail within logs significantly influences the accuracy of retrieving relevant context. Similarly, the way questions are formulated by users affects retrieval quality. For example, prefacing identifiers with an ``ID'' tag, as in changing \textit{Navigation to waypoint 5 has succeeded} to \textit{Navigation to the waypoint with ID:5 has succeeded}, has enhanced retrieval outcomes. This highlights the importance of both log verbosity and precise question formulation.

\subsection{Experimentation}

To evaluate our proposed system, we conducted an experiment focusing on a navigation task. Given the complexity of autonomous robot behavior and the potential need for self-explanatory capabilities across various skills, our system is designed to be skill-agnostic. However, for our initial assessment, we chose to concentrate on a particular task.

The primary aim is to assess the quality of explanations provided by the LLM. Emphasizing the significance of a real-world context for our experiment, we based our testing on an actual European Robotics League (ERL) navigation benchmark. This section describes the experimental setup.

\bmhead{ERL Navigation Benchmark}

Choosing an ERL test reflects our desire to simulate a real-world environment where the robot might face unforeseen challenges. This benchmark tests the robot's ability to navigate an apartment setting, utilizing the Leon@home testbed.

In the ERL benchmark, the robot begins at a set starting point and must navigate to three sequential waypoints. The test is completed when the robot arrives at the final waypoint.

To adapt the original ERL test to our experimental needs, we made certain adjustments. Firstly, instead of receiving waypoints through a topic as in the standard ERL test, waypoints are generated within the waypoint\_navigation node for our experiment. Secondly, to ensure uniform data for all test runs, waypoints are predetermined rather than randomly generated. This setup directs the robot to navigate to waypoints 9, 6, and 7, in that order.

\bmhead{Experiment design}

\begin{figure*}[h]%
\centering
\includegraphics[width=0.9\textwidth]{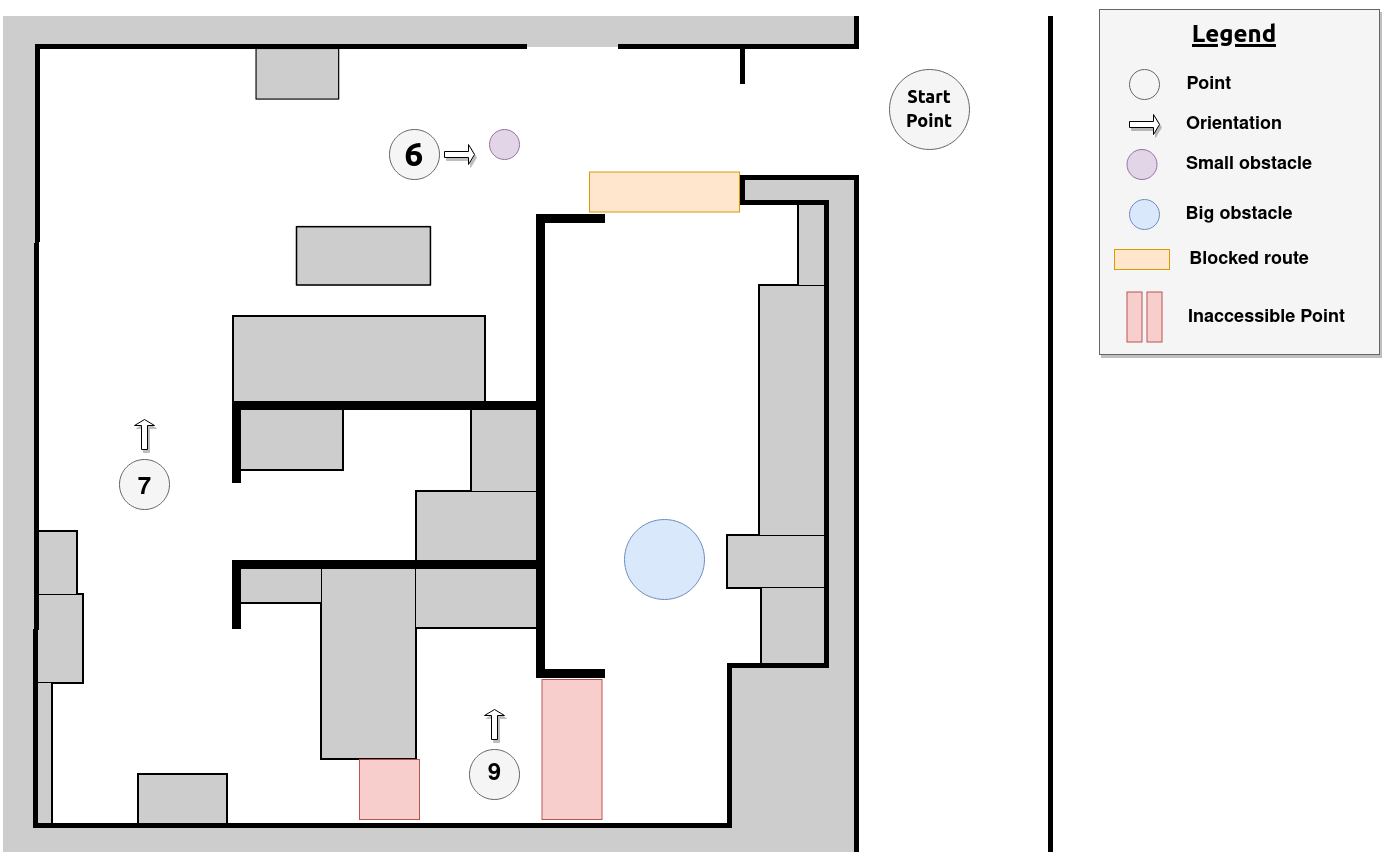}
\caption{Testbed map for the experiment}\label{fig:apartment}
\end{figure*}

The map in Figure \ref{fig:apartment} showcases the testbed layout, including the designated robot route and obstacles.

The experiment consisted of various runs (R) or trials as follows:

\begin{description}
\item[\textbf{R1}] Navigation without any obstacles.
\item[\textbf{R2}] Navigation encountering small obstacles.
\item[\textbf{R3}] Navigation with large obstacles in the path.
\item[\textbf{R4}] Navigation facing a complete route blockage.
\item[\textbf{R5}] Navigation with a blocked route where replanning is not an option.
\end{description}

\subsection{Evaluation}

This section details our approach to evaluating the conducted experiment. We defined a set of metrics to analyze the explanation system's performance and designed a questionnaire to assess the quality of the explanations generated.

\bmhead{Metrics}

The metrics derived from the experiment aim to quantify the volume of information forming the system's context and the duration needed to process this information and produce an explanation.

\begin{itemize}
    \item \textbf{Task execution time (s)}.
    \item \textbf{Number of messages published in the final rosbag.}
    \item \textbf{Number of messages published in the \textit{/rosout} topic.}
    \item \textbf{Number of \textit{/rousout} messages processed.}
    \item \textbf{Log processing time (s).}
    \item \textbf{Question generation time (s).}
\end{itemize}

After completing the test and compiling the logs into the vector database, the system is ready to address questions about the navigation task. We formulated the following User Questions (UQ) for this purpose:

\begin{description}
    \item [\textbf{UQ1}] How many waypoints were received during the navigation task?
    \item [\textbf{UQ2}] Which were the IDs of the waypoints received during the navigation task?
    \item [\textbf{UQ3}] Were all the waypoints received successfully reached?
    \item [\textbf{UQ4}] What happened during navigation to waypoint with ID X?
    \item [\textbf{UQ5}] Why was the route replanned during navigation to waypoint with ID X?
    \item [\textbf{UQ6}] Have any relevant events occurred during navigation?
    \item [\textbf{UQ7}] What is the task that the robot had to perform?
    \item [\textbf{UQ8}] Did the robot avoid any obstacles during the navigation?
\end{description}

These questions are designed to clarify events during navigation, ranging from general inquiries to specific ones focused on particular waypoints. For questions about specific waypoints, the placeholder "X" will be replaced with "ID 6" or "ID 9" in actual queries, corresponding to obstacle encounters or route changes.

\bmhead{Evaluation questionnaire}

The system allows users to query an LLM to explain the robot's behavior. To assess the quality of these explanations, we developed a questionnaire targeting \textbf{technical users} specifically, recognizing that explanation acceptability can vary with the user's expertise. The questionnaire is inspired by scales from \cite{alves2015empathic} and \cite{bhattacherjee2004understanding}, focusing on expectation and satisfaction in HRI contexts.

Questionnaire items are categorized into two groups, shown in Table \ref{tab:itemsScale}, assessing explanation quality and satisfaction.

\begin{table*}[h]
\centering
\rowcolors{2}{white}{lightgray}
\caption{Validation scale items. Each Evaluation Question (EQ) has an ID, a class, and an item description.}\label{tab:itemsScale}
\begin{tabular*}{\textwidth}{p{0.7cm}p{3.8cm}p{10.2cm}}
\toprule
ID & Class & Item\\
\midrule
\(EQ_{1}\) & Explanation Quality & I consider the explanation to be unnecessarily long\\
\(EQ_{2}\) & Explanation Quality & The explanation is written in clear and understandable language\\
\(EQ_{3}\) & Explanation Quality & I believe that the explanation is consistent with the robot's behavior shown in the logs\\
\(EQ_{4}\) & Explanation Satisfaction & The explanation has given me a better understanding of what happened during robot task\\
\(EQ_{5}\) & Explanation Satisfaction & I consider that the explanation evidences the behavior of the robot\\
\(EQ_{6}\) & Explanation Satisfaction & I consider that the explanation is sufficiently reliable and accurate to understand the behavior of the robot\\
\botrule
\end{tabular*}
\end{table*}

Responses are gathered using a 5-point Likert scale, as used in \cite{likert1932technique}, detailed in Table \ref{tab:likertScale}.

\begin{table*}[h!]
\centering
\caption{Ranges of the Likert scale used.}\label{tab:likertScale}
    \begin{tabular*}{\textwidth}{ccccc}
        \toprule
         Strongly Disagree (SD) & Disagree (D) & Neutral (N)	& Agree (A)	& Strongly Agree (SA)\\
        \midrule
        1 & 2 & 3 & 4 & 5\\
         \bottomrule
    \end{tabular*}
\end{table*}

The questionnaire is designed alongside the logs from the robot's operation to validate the LLM's responses. To streamline the evaluation, we selected run 4 for its involvement in path replanning -- a significant navigational event. 

\section{Results}\label{sec:results}

This section outlines the results obtained from our experiment. Initially, we will discuss the navigation and log collection outcomes, highlighted by the additional logs introduced during navigation to enhance verbosity, which are not directly linked to Nav2 (see Listing \ref{lst:ejemploLogs}).

\lstset{frame=none, basicstyle=\footnotesize\ttfamily, backgroundcolor=\color{lightgray},  linewidth=0.95\columnwidth, language=pascal, breaklines=true, caption="Example of enhanced log.", label=lst:ejemploLogs}
\begin{lstlisting}
A list of waypoints has been received
The waypoints received are: 9 6 7
Waypoint with ID: 9 has been received
Navigating to the waypoint with ID:9
Navigation to the waypoint with ID: 9 is in progress.
Navigation to the waypoint with ID: 9 has succeeded.
Waiting for a new waypoint...
Waypoint with ID: 6 has been received
Navigating to the waypoint with ID:6
Navigation to the waypoint with ID: 6 is in progress.
Navigation to the waypoint with ID: 6 has succeeded.
Waiting for a new waypoint...
Waypoint with ID: 7 has been received
Navigating to the waypoint with ID:7
Navigation to the waypoint with ID: 7 is in progress.
Obstacle detected during navigation to waypoint with ID:7 - Distance to the point increased from: 0.07 meters to 0.10 meters
Navigation to the waypoint with ID: 7 has succeeded.
Waiting for a new waypoint...
All the waypoints received have been reached. Navigation task completed.
\end{lstlisting}

As mentioned in section \ref{sub:desafios}, the issue of bad retrievals is critical. The structure of logs, as shown above, illustrates the significance of this problem. Listing \ref{lst:badretrieval} demonstrates how repeated messages can saturate the retrieval process, introducing excess noise and leading to irrelevant responses. Conversely, listing \ref{lst:goodretrieval} displays improved retrieval outcomes after implementing the solutions proposed in the Challenges section.

\lstset{frame=none, basicstyle=\footnotesize\ttfamily, backgroundcolor=\color{lightgray},  linewidth=0.95\columnwidth, language=pascal, breaklines=true, caption=Relevant messages retrieval with repeated messages., label=lst:badretrieval}
\begin{lstlisting}
Prompt after formatting:
You are an explainability AI for 
    autonomous robots.
You are designed to provide 
    explications about the robot 
    logs, providing only factual 
    information.
You have to interpret the logs 
    that have been generated 
    during a run.
You should not be overly chatty.
Relevant Logs are below.:
Passing new path to controller.
Passing new path to controller.
Passing new path to controller.
Passing new path to controller.
Passing new path to controller.
Passing new path to controller.
Passing new path to controller.
Passing new path to controller.
Passing new path to controller.
Passing new path to controller.
Passing new path to controller.
Passing new path to controller.
Passing new path to controller.
Passing new path to controller.
Passing new path to controller.
Passing new path to controller.
Passing new path to controller.
Passing new path to controller.
Passing new path to controller.
Passing new path to controller.
\end{lstlisting}

\lstset{frame=none, basicstyle=\footnotesize\ttfamily, backgroundcolor=\color{lightgray},  linewidth=0.95\columnwidth, language=pascal, breaklines=true, caption=Relevant messages retrieval without repeated messages., label=lst:goodretrieval}
\begin{lstlisting}
Prompt after formatting:
You are an explainability AI for autonomous robots.
You are designed to provide explanations about the robot logs, providing only factual information.
You have to interpret the logs that have been generated during a run.
You should not be overly chatty.
Relevant logs are below.:
Received a goal, begin computing control effort.
Received a goal, begin computing control effort.
Received a goal, begin computing control effort.
No goal checker was specified in parameter 'current_goal_checker'. Server will use only plugin loaded general_goal_checker. This warning will appear once.
Waiting for a new waypoint...
Waiting for a new waypoint...
Waiting for a new waypoint...
Navigation to the waypoint with ID: 5 has succeeded.
Navigation to the waypoint with ID: 8 has succeeded.
All the waypoints received have been reached. Navigation task completed.
Navigation to the waypoint with ID: 8 is in progress.
Navigating to the waypoint with ID:5
Navigation to the waypoint with ID: 5 is in progress.
Navigating to the waypoint with ID:8
Navigating to the waypoint with ID:3 A list of waypoints has been received
Begin navigating from current location (0.00, 10.00) to (-10.42, -33.74)
Begin navigating from current location (4.63, 19.74) to (-0.28, -27.48)
Begin navigating from current location (-10.28, -33.65) to (4.55, 19.94)
\end{lstlisting}

The chosen parameters for llama\_ros configuration are detailed in Tables \ref{tab:confParams} (configuration parameters) and \ref{tab:sampParams} (sampling parameters), optimized based on system resources.

\begin{table}[h!]
\centering
\rowcolors{2}{white}{lightgray}
\caption{Configuration parameters for llama\_ros.}\label{tab:confParams}
  \begin{tabular}{p{3cm} p{3cm}}
    \toprule
    Parameter Name & Parameter Value\\
    \midrule
    n\_threads & 4\\
    n\_ctx & 4096\\
    n\_batch & 256\\ 
    n\_gpu\_layers & 33\\
    \bottomrule
  \end{tabular}
\end{table}

\begin{table}[h!]
\centering
\rowcolors{2}{white}{lightgray}
\caption{Sampling parameters employed to generate responses with llama\_ros.}\label{tab:sampParams}
  \begin{tabular}{p{3cm} p{3cm}}
    \toprule
    Parameter Name & Parameter Value\\
    \midrule
    n\_prev & 64\\
    top\_k & 40\\
    top\_p & 0.95f\\
    temp & 0.00f\\
    penalty\_last\_n & 64\\
  \bottomrule
  \end{tabular}
\end{table}

The experiment's data are divided into three categories: ERL navigation benchmark results, explanation generation outcomes, and questionnaire feedback.

\bmhead{ERL navigation benchmark results}

Table \ref{tab:ProcessingData} summarizes the metrics from each navigation test, including the number of logs recorded and the processing time required by the system.

\begin{table*}[h!]
\centering
\rowcolors{2}{white}{lightgray}
\caption{Amount of logs collected. The number of logs published in rosout and the number of logs processed during the five executions (Runs). Execution time of the run and processing time for total logs of each run.}\label{tab:ProcessingData}
    \begin{tabular*}{\textwidth}{p{1.5cm} p{1.5cm} p{2.7cm} p{2.7cm} p{3.5cm} p{1.5cm}}
         \toprule
         Run & Execution Time(s) & Rosbag entries & Total Logs in Rosout & Embeddings Processed & Processing Time(s)\\
        \midrule
          1 & 162.2 & 36161 & 168 & 62 & 16.824\\
          2 & 150.2 & 35806 & 127 & 119 & 30.230\\
          3 & 154.3 & 34517 & 181 & 76 & 20.687\\
          4 & 190.23 & 43996 & 219 & 75 & 20.316\\
          5 & 218.05 & 49641 & 342 & 148 & 96.223\\
         \bottomrule
    \end{tabular*}
\end{table*}

\bmhead{Explanation generation results}

After establishing the necessary context, the system is ready to generate explanations. Results are organized into question-answer pairs for each execution. For shortness, we only include explanations from execution 4 here (see Table \ref{tab:expExecution4}), with further details in appendix \ref{secA1}.

\begin{table*}[]
\centering
\small
\rowcolors{2}{white}{lightgray}
\caption{Explanations obtained in run 4.}\label{tab:expExecution4}
\begin{tabular}{p{.3\textwidth} p{.65\textwidth}}
\toprule
\textbf{User Question} & \textbf{LLM's Response} \\
\midrule
How many waypoints were received during the navigation task? & During the navigation task, 3 waypoints were received. They are with IDs: 9, 6, and 7. All of them have been reached successfully\\

What were the IDs of the waypoints received during the navigation task? & The waypoints received during the navigation task had IDs: 9, 6, and 7\\

Were all the waypoints received successfully reached? & Yes, all the waypoints received have been reached. Navigation task completed\\

What happened during navigation to waypoint with ID 6? & During navigation to waypoint with ID 6, initially there was an issue with the path as it was reported as invalid and empty. This led to receiving a goal multiple times for computing control effort. Eventually, the robot reached the goal associated with waypoint 6, and the navigation to this waypoint succeeded. After reaching the goal, the robot waited for a new waypoint\\

Why was the route replanned during navigation to waypoint with ID 6? & The route was replanned during navigation to waypoint with ID 6 due to an Obstacle detected. The distance to a certain point increased from 5.78 meters to 14.65 meters, which indicates that the robot encountered an obstacle in its path and needed to adjust its course accordingly to reach the goal while avoiding the obstacle\\

Have any relevant events occurred during navigation? & Yes, several relevant events occurred during navigation. These include reaching goals, goal successes, and waypoint navigation completions. Additionally, there were instances of encountering invalid paths and waiting for new waypoints. The navigation task was eventually completed with all waypoints reached\\

What is the task that the robot had to perform? & The task that the robot had to perform was navigating to various waypoints with specific IDs and reaching those goals. It involved computing control efforts, moving towards the goal locations, and waiting for new waypoint instructions. The navigation process continued until all the given waypoints were reached, completing the overall navigation task\\

Did the robot avoid any obstacle during the navigation? & The provided logs do not mention any specific details about encountering or avoiding obstacles during the navigation. Therefore, it is impossible to determine if the robot avoided any obstacles based on this information alone\\
\botrule
\end{tabular}
\end{table*}

Table \ref{tab:tiempoRespuestaReal} shows the LLM response times for user questions across all executions, highlighting the efficiency of the explanation generation process.

\begin{table*}[h!]
\centering
\rowcolors{2}{white}{lightgray}
\caption{Average LLM response time (s) for User Questions during the five executions (Runs).}\label{tab:tiempoRespuestaReal}
    \begin{tabular*}{\textwidth}{p{1.8cm}p{1.3cm}p{1.3cm}p{1.3cm}p{1.3cm}p{1.3cm}p{1.3cm}p{1.3cm}p{1.3cm}}
         \toprule
         & \(UQ_{1}\) & \(UQ_{2}\) & \(UQ_{3}\) & \(UQ_{4}\) & \(UQ_{5}\) & \(UQ_{6}\) & \(UQ_{7}\) & \(UQ_{8}\)\\
        \midrule
         Run 1 & 4.92 & 4.58 & 4.16 & 5.70 & 7.13 & 5.94 & 5.67 & 6.31\\
         Run 2 & 4.92 & 4.51 & 4.06 & 5.65 & 5.38 & 5.56 & 5.78 & 6.50\\
         Run 3 & 4.91 & 4.68 & 4.37 & 5.96 & 6.13 & 5.88 & 5.56 & 6.50\\
         Run 4 & 4.69 & 4.43 & 4.04 & 6.30 & 6.65 & 4.99 & 5.89 & 4.80\\
         Run 5 & 4.78 & 4.58 & 6.41 & 6.60 & 8.28 & 5.81 & 4.81 & 5.39\\
         Mean & 4.84 & 4.55 & 4.61 & 6.04 & 6.71 & 5.63 & 5.54 & 5.9\\
         \botrule
         \end{tabular*}
\end{table*}

\bmhead{Validation questionnaire results}

\begin{figure}[h!]%
\centering
\includegraphics[width=0.5\textwidth]{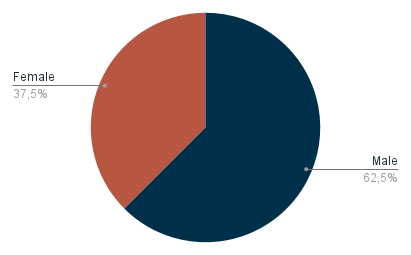}
\caption{Self-Identification Among Participants}\label{fig:genrePlot}
\end{figure}

\begin{figure}[h!]%
\centering
\includegraphics[width=0.5\textwidth]{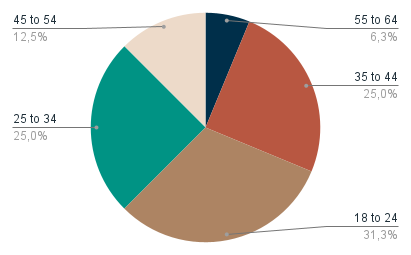}
\caption{Age Distribution of Participants}\label{fig:agePlot}
\end{figure}

\begin{figure}[h!]%
\centering
\includegraphics[width=0.5\textwidth]{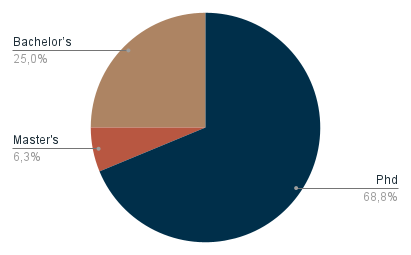}
\caption{Highest Education Levels Among Survey Participants}\label{fig:studiesPlot}
\end{figure}

Sixteen participants, including 10 men and 6 women with varying educational backgrounds in robotics, participated in the questionnaire. Figures \ref{fig:genrePlot}, \ref{fig:agePlot}, and \ref{fig:studiesPlot} display demographic data.

\begin{figure*}[h!]%
\centering
\includegraphics[width=0.97\textwidth]{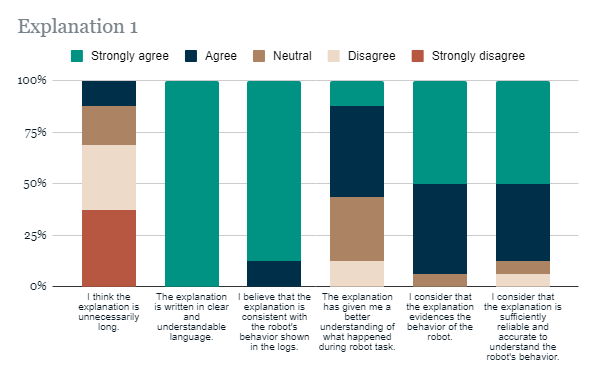}
\caption{Participant's evaluation to explanation 1 for run 4}\label{fig:explanationExample}
\end{figure*}

The feedback on LLM explanations is presented using stacked box plots. These plots showcase participant perceptions for each evaluated explanation. Figure \ref{fig:explanationExample} shows an example of the plots generated from the questionnaire results. The remaining figures are gathered in appendix \ref{secB1}.

\section{Discussion}\label{sec:discussion}

This section delves into the outcomes observed during the experimental phase, structured into distinct subsections for clarity.

\bmhead{Processing time}

A close examination of log processing times in Table \ref{tab:ProcessingData} reveals generally swift durations, typically ranging between 20 to 30 seconds, except for Run 5, which extends to about 90 seconds. This anomaly aligns with its prolonged execution time (218s) and a larger batch of processed embeddings (148 logs). The delay in Run 5 stems from the robot's replanning challenges, escalating the log count significantly.

The processing duration for embeddings hinges on two factors: the volume and complexity of the logs generated. Extended tasks naturally lead to a higher production of logs, demanding more extensive processing. Complex or lengthy log messages further prolong the necessary processing time.

As highlighted in section \ref{sub:desafios}, achieving real-time responses poses a challenge, primarily due to the time required for log processing. Experimental data corroborates this, with response generation to user queries taking around 5 to 7 seconds, varying by the complexity and length of the response (refer to table \ref{tab:tiempoRespuestaReal}). Although current processing times are manageable, real-time explanation delivery remains unattainable with the current setup. Future advancements in hardware and model efficiency could potentially ameliorate these constraints.

\bmhead{Explanation obtained}

This segment evaluates the quality of the explanations generated. Detailed in \ref{secA1} (Runs 1-3, and 5) and Table \ref{tab:expExecution4} (run 4), the explanations reflect the system's initial expectations. Yet, an internal assessment alone isn't sufficient to ensure objectivity. Thus, insights from the questionnaire supplement this discussion, offering a broader perspective on the explanations' efficacy.

Questions 1, 2, 3, and 7 were consistently answered with precision across all trials. These queries, which demand specific information (like waypoint numbers or task details), benefit from the logs' high verbosity, minimizing the need for inference. Hence, their responses largely hinge on the retrieval process's accuracy.

Particularly remarkable is the response to question 3 in Run 5, a complex scenario where navigation was prematurely halted without additional explanatory logs. Yet, the model adeptly concluded the incomplete waypoint navigation.

However, responses to question 8 were uniformly incorrect across all runs. This discrepancy arises from two main issues: the misinterpretation of replanning logs and the retrieval process's shortcomings. Notably, the absence of explicit mentions concerning obstacle avoidance or replanning in Runs 2, 3, and 4 underscores the model's overlook of minor replanning logs.

Question 5's focus on replanning should theoretically yield precise retrieval outcomes. Nevertheless, assumptions embedded within the question and noise from navigation logs may skew the context, affecting accuracy.

The less definitive queries allowed the model more interpretive leeway. For example, question 4 received accurate responses in most cases, with the explanation for Run 5 highlighting the model's capability to deduce repeated navigation attempts without success.

Evaluating question 6 proves more complex, given its abstract formulation. While the model's focus on successful waypoint navigation is pertinent, a question rephrase might better align expectations with outcomes.

\subsection{Questionnaire results}

The questionnaire analysis is segmented by the Evaluation Questions.

\subparagraph{I consider the explanation to be unnecessarily long (EQ1)} Most participants found the explanations appropriately concise, except for responses to explanations 5, 7, and 8, where a desire for brevity was noted. This suggests a potential for reducing verbosity without a strong demand for it.

\subparagraph{The explanation is written in clear and understandable language (EQ2)} Evaluating linguistic clarity, the feedback underscores LLMs' proficiency in natural language processing, with explanation 4 receiving a nuanced view due to its ambiguous clarity.

\subparagraph{I believe that the explanation is consistent with the robot’s behavior shown in the logs (EQ3)} Generally, participants affirmed the explanations' consistency with observed robot behaviors, except for explanations 4 and 8, which inadequately reflected the events, notably missing obstacle detection details.

\subparagraph{The explanation has given me a better understanding of what happened during the robot task (EQ4)} Disagreement was evident in explanations 4, 6, and 8, with explanation 8 particularly criticized for inaccuracies, suggesting a misalignment with the actual events.

\subparagraph{I consider that the explanation evidences the behavior of the robot (EQ5)} Most participants believe that explanation 4 and, notably, explanation 8 do not adequately showcase the behavior of the robot. This observation becomes evident when we note that both explanations omit any mention of obstacle detection.

\subparagraph{I consider that the explanation is sufficiently reliable and accurate to understand the behavior of the robot (EQ6)} Once again, explanation 8 was deemed unreliable, highlighting the importance of accuracy and reality reflection in responses.

\section{Conclusions}\label{sec:conclusions}

Explaining the behavior of autonomous systems presents a considerable challenge. While humans might find it straightforward to explain specific events, the task becomes significantly more complex and abstract when assigned to robots. In this research, we've advanced the field of explainability by creating a system that leverages LLM technology to interpret logs from an autonomous robot. This system is designed to construct explanations for the robot's actions based on log data.

Our experiment explored LLMs' capability to decipher technical data. Although the responses are not yet flawless, the findings underscore the potential of machine learning models to enhance explainability. The developed system, alongside the experiment, demonstrates that LLMs can act as efficient generators of explanations and interpreters of logs (\textbf{RQ1}). Furthermore, this study highlights the critical role of log verbosity (\textbf{RQ2}) and the precision of retrieval methods. Together, these elements are crucial for producing high-quality explanations.

Remarkably, our system's development utilized various LLM models, reflecting the rapid evolution within LLM technology. This evolution means that updates to the system's model could significantly impact both performance and the quality of explanations, necessitating periodic model refreshes.

Looking ahead, we plan to refine the system to tailor explanations to the user's expertise level, aiming for variable detail in responses. We also intend to optimize the retrieval system to reduce log processing and explanation generation times, improve explanation context, and extend the proof of concept to tasks beyond navigation. The need to enhance event tracing during navigation became evident, as this aspect influenced several obtained explanations. Insights from analyzing the questionnaire were invaluable, indicating that rephrasing some questions could lead to more precise explanations. This underscores the importance of carefully crafting user queries to improve retrieval outcomes.

\backmatter

\bmhead{Acknowledgments}

This research has been partially funded by the Recovery, Transformation, and Resilience Plan, financed by the European Union (Next Generation) thanks to the TESCAC project (Traceability and Explainability in Autonomous Cystems for improved Cybersecurity) granted by INCIBE to the University of León, and by grant PID2021-126592OB-C21 funded by MCIN/AEI/10.13039/501100011033
EDMAR (Explainable Decision Making in Autonomous Robots) project, PID2021-126592OB-C21funded by MCIN/AEI/10.13039/501100011033 and by ERDF ''A way of making Europe''.

\section*{Declarations}

\begin{itemize}
\item \textbf{Conflict of interest} The author declares no conflict of interest
\item \textbf{Ethics approval} The authors confirm that they have complied with the publication ethics and state that this work is original and has not been used for publication anywhere before.
\item \textbf{Consent to participate} All questionnaire participants have given their consent to participate. The study is anonymous without collecting personal data.
\item \textbf{Consent for publication} All questionnaire participants have given their consent to publish. The study is anonymous without collecting personal data.
\item \textbf{Availability of data and materials} Source code available in: \url{https://github.com/Dsobh/explainable_ROS}. Rosbag's dataset is available in: \url{https://zenodo.org/records/10518775}
\end{itemize}


\begin{appendices}

\section{Supplementary materials}

To ensure the main text of the paper remains organized and is not overloaded with data, this section compiles all essential information regarding the experimental results in the following tables and figures. Also, this section includes the prompt template used.

\subsection{Prompt template}\label{secA0}

\lstset{frame=none, basicstyle=\footnotesize\ttfamily, backgroundcolor=\color{lightgray},  linewidth=0.95\columnwidth, language=pascal, breaklines=true, caption=Prompt template used to generate explanations., label=lst:template}
\begin{lstlisting}
"<|im_start|>system\n"
"You are an explainability AI for autonomous robots.\n"
"You are designed to provide explanations about the robot logs, providing only factual information.\n"
"You have to interpret the logs that have been generated during a run.\n"
"You should not be overly chatty.\n"

"Relevant logs are below.:\n"
"{logs}<|im_end|>\n"

"<|im_start|>user\n"
"Given the context information and no prior knowledge, answer the query: {question}<|im_end|>\n"
"<|im_start|>assistant\n"
\end{lstlisting}

\subsection{Explanation obtained during executions}\label{secA1}

This section gathers the results obtained during the different runs. First, Table \ref{tab:expExecution1}, contains the LLM's responses corresponding to run 1, which consists of a navigation task without obstacles. The second run performed during the experiment provides the explanations documented in Table \ref{tab:expExecution2}. Table \ref{tab:expExecution3} shows the results obtained for Execution 3. Both runs (2 and 3) include navigation with obstacle detection issues. Finally, Table \ref{tab:expExecution5}, contains the explanation generated for execution 5, wherein all possible routes available to complete the task are locked.

\begin{table*}[h!]
\centering
\rowcolors{2}{white}{lightgray}
\caption{Explanations obtained for execution 1.}\label{tab:expExecution1}
\begin{tabular}{p{.3\textwidth} p{.65\textwidth}}
\toprule
\textbf{User Question} & \textbf{LLM's Response} \\
\midrule
How many waypoints were received during the navigation task? & During the navigation task, 3 waypoints were received. They are with IDs: 9, 6, and 7. All of them have been reached successfully\\

What were the IDs of the waypoints received during the navigation task? & The waypoints received during the navigation task had IDs: 9, 6, and 7\\

Were all the waypoints received successfully reached? & Yes, all the waypoints received have been reached. Navigation task completed\\

What happened during navigation to waypoint with ID 6? & During navigation to the waypoint with ID 6, the robot received a goal and began computing control effort. It successfully reached the goal and navigated to the waypoint with ID: 6, which then had succeeded. After reaching this waypoint, it waited for a new waypoint before proceeding further\\

Why was the route replanned during navigation to waypoint with ID 6? & The route was replanned during navigation to waypoint with ID 6 due to an obstacle detection. The distance to a certain point increased from 0.12 meters to 0.16 meters, which indicates that the robot encountered an unexpected obstacle in its path. This required adjusting the planned route to navigate around or through the obstacle and continue towards the waypoint with ID 6\\

Have any relevant events occurred during navigation? & Yes, several relevant events occurred during navigation. The robot navigated to different waypoints (9, 6, and 7), reached goals at each location, and eventually completed the navigation task by reaching all the given waypoints. Additionally, it encountered a warning about the goal checker parameter not being specified but continued functioning without issues\\

What is the task that the robot had to perform? & The task that the robot had to perform was navigating to a series of waypoints (9, 6, and 7) and reaching them one by one. After completing all the waypoints, it waited for new instructions or waypoints. Once all the given waypoints were reached, the navigation task was considered completed\\

Did the robot avoid any obstacle during the navigation? & Yes, the robot encountered obstacles during its navigation. It detected an obstacle while navigating to waypoint with ID: 6 and another one while navigating to waypoint with ID: 7. The distance to the points increased due to these obstacles. However, it managed to reach all the waypoints despite the obstacles\\
\botrule
\end{tabular}
\end{table*}

\begin{table*}[h!]
\centering
\rowcolors{2}{white}{lightgray}
\small
\caption{Explanations obtained for execution 2.}\label{tab:expExecution2}
\begin{tabular}{p{.3\textwidth} p{.65\textwidth}}
\toprule
\textbf{User Question} & \textbf{LLM's Response} \\
\midrule
How many waypoints were received during the navigation task? & During the navigation task, 3 waypoints were received. They are with IDs: 9, 6, and 7. All of them have been reached successfully\\

What were the IDs of the waypoints received during the navigation task? & The IDs of the waypoints received during the navigation task were 9, 6, and 7\\

Were all the waypoints received successfully reached? & Yes, all the waypoints received have been reached. Navigation task completed\\

What happened during navigation to waypoint with ID 6? & During navigation to waypoint with ID 6, the robot received a goal, began computing control effort, navigated to the waypoint, reached the goal, and successfully completed the task. After that, it waited for a new waypoint\\

Why was the route replanned during navigation to waypoint with ID 6? & There is no mention of a route being replanned during navigation to waypoint with ID 6 in the provided logs. The logs only show successful navigation and reaching goals for each mentioned waypoint. No specific reason for replanning was given\\

Have any relevant events occurred during navigation? & Yes, several relevant events occurred during navigation. The robot navigated to different waypoints (9, 6, and 7), reached goals at each of these waypoints, and eventually completed the navigation task by reaching all the given waypoints. Additionally, it recorded logs throughout the process\\

What is the task that the robot had to perform? & The task that the robot had to perform was navigating to various waypoints with different IDs (9, 6, and 7) and reaching them. After completing all the waypoint navigation tasks, it waited for new waypoints and once there were no more waypoints, it concluded the navigation task as completed\\

Did the robot avoid any obstacle during the navigation? & During the navigation, an obstacle was detected while navigating to waypoint with ID: 7. The distance to the point increased from 0.11 meters to 0.16 meters. However, it is not explicitly mentioned if the robot successfully avoided the obstacle or not. It only managed to reach the goal and complete the navigation task\\
\botrule
\end{tabular}
\end{table*}

\begin{table*}[h!]
\centering
\rowcolors{2}{white}{lightgray}
\small
\caption{Explanations obtained for execution 3.}\label{tab:expExecution3}
\begin{tabular}{p{.3\textwidth} p{.65\textwidth}}
\toprule
\textbf{User Question} & \textbf{LLM's Response} \\
\midrule
How many waypoints were received during the navigation task? & During the navigation task, 3 waypoints were received. They are with IDs: 9, 6, and 7. All of them have been reached successfully\\

What were the IDs of the waypoints received during the navigation task? & The waypoints received during the navigation task had IDs: 9, 6, 7\\

Were all the waypoints received successfully reached? & Yes, all the waypoints received have been successfully reached. The navigation task was completed as mentioned in the log "Navigation task completed\\

What happened during navigation to waypoint with ID 6? & During the given log, there is no specific mention of navigation to a waypoint with ID 6. The robot navigated to waypoints with IDs 9 and 7, but not explicitly for ID 6. Therefore, we cannot provide information about what happened during navigation to waypoint with ID 6 based on these logs alone\\

Why was the route replanned during navigation to waypoint with ID 6? & There is no mention of a route being replanned or any issue related to waypoint with ID 6 in the provided logs. The logs only discuss navigation and reaching goals for different waypoints, but there's no specific information about a change in route or reason for replanning during navigation to waypoint 6\\

Have any relevant events occurred during navigation? & Yes, during navigation, several waypoints were reached, including reaching goals at each waypoint. The robot navigated to waypoints with IDs 9, 6, and 7, and eventually completed the navigation task by reaching all the given waypoints. Additionally, a warning about the goal checker was mentioned but did not affect the navigation process\\

What is the task that the robot had to perform? & The task that the robot had to perform was navigating to different waypoints with specific IDs and reaching those goals. It involved listening for topics, recording, computing control efforts, and waiting for new waypoints as needed. The navigation process continued until all the waypoints were reached, completing the overall navigation task\\

Did the robot avoid any obstacle during the navigation? & Yes, the robot encountered an obstacle during the navigation to waypoint with ID: 9 and another one during the navigation to waypoint with ID: 7. In both cases, it detected an increase in distance due to the presence of obstacles. However, it managed to reach its goals despite these obstacles\\
\botrule
\end{tabular}
\end{table*}

\begin{table*}[h!]
\centering
\rowcolors{2}{white}{lightgray}
\small
\caption{Explanations obtained for execution 5.}\label{tab:expExecution5}
\begin{tabular}{p{.3\textwidth} p{.65\textwidth}}
\toprule
\textbf{User Question} & \textbf{LLM's Response} \\
\midrule
How many waypoints were received during the navigation task? & During the navigation task, 3 waypoints were received. They are with IDs: 9, 6, and 7\\

What were the IDs of the waypoints received during the navigation task? & The waypoints received during the navigation task had IDs: 9, 6, and 7\\

Were all the waypoints received successfully reached? & Not all the waypoints were received successfully reached. The log shows that navigation to the waypoint with ID: 9 has aborted, indicating it was not completed successfully. Additionally, there are instances of "Invalid path, Path is empty" which also indicate unsuccessful attempts at reaching certain points. Other waypoints might have been reached, but these specific issues prevented successful completion for some waypoints\\

What happened during navigation to waypoint with ID 6? &  During the navigation to waypoint with ID 9, several attempts were made. The robot encountered issues such as invalid paths, empty paths, and failed plan creation. Eventually, the navigation aborted, and a new waypoint was awaited. Additionally, there were some warnings and goal-related messages throughout the process\\

Why was the route replanned during navigation to waypoint with ID 6? & The route had to be replanned during navigation to waypoint with ID 9 due to obstacle detections and issues encountered along the path. Obstacles were detected multiple times, causing an increase in distance to the target point. Additionally, GridBased planning failed several times, resulting in invalid paths or empty paths. These factors led to the need for replanning the route to navigate around the obstacles and find a suitable path towards the waypoint with ID 9. Eventually, navigation to this waypoint was aborted due to ongoing issues\\

Have any relevant events occurred during navigation? & Yes, several relevant events occurred during navigation. These include receiving waypoints, navigating to specific waypoints, encountering invalid paths or empty paths, goal cancellation, and a critical failure with the server shutting down. Additionally, there were instances of recording and subscribing to topics like '/scan' and '/map'\\

What is the task that the robot had to perform? & The task that the robot had to perform was navigating to a waypoint with ID: 9. However, it faced issues due to invalid paths and eventually aborted the navigation process\\

Did the robot avoid any obstacle during the navigation? & Based on the provided logs, there is no mention of encountering or avoiding any obstacles during the navigation process. The focus was mainly on reaching waypoints and dealing with planning issues. However, without more information, it's impossible to confirm if an obstacle was indeed avoided\\
\botrule
\end{tabular}
\end{table*}

\subsection{Questionnarie results}\label{secB1}

The present subsection shows the questionnaire results obtained for questions (2-5) asked to participants. The explanation evaluated in the questionnaire corresponds to execution 4 of the experiment.

\begin{figure*}[h!]%
\centering
\includegraphics[width=0.97\textwidth]{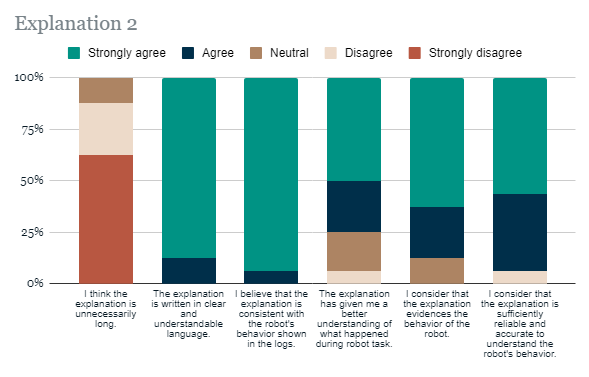}
\caption{Participant's evaluation to explanation 2 for run 4}\label{fig:explanation1}
\end{figure*}

\begin{figure*}[h!]%
\centering
\includegraphics[width=0.97\textwidth]{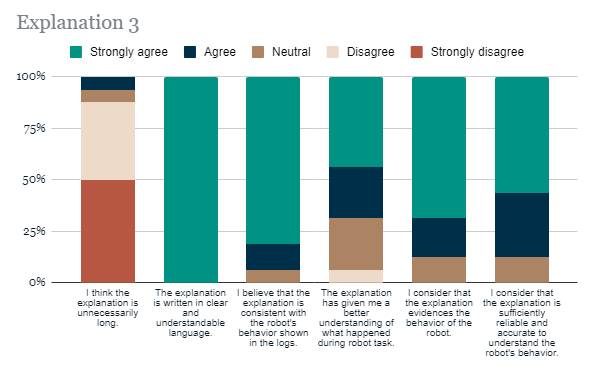}
\caption{Participant's evaluation to explanation 3 for run 4}\label{fig:explanation1}
\end{figure*}

\begin{figure*}[h!]%
\centering
\includegraphics[width=0.97\textwidth]{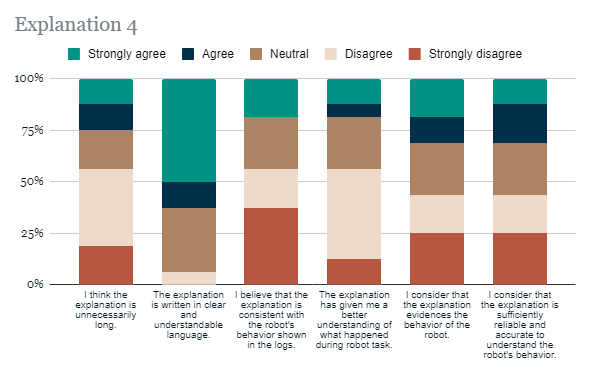}
\caption{Participant's evaluation to explanation 4 for run 4}\label{fig:explanation1}
\end{figure*}

\begin{figure*}[h!]%
\centering
\includegraphics[width=0.97\textwidth]{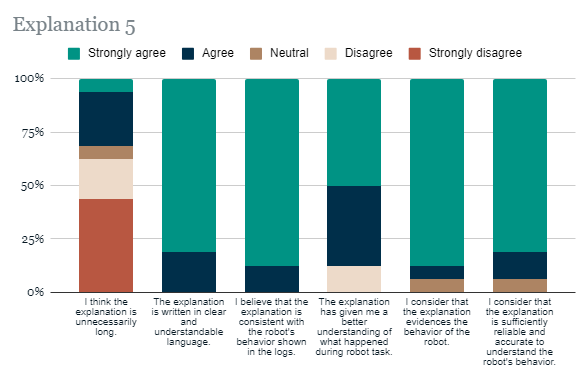}
\caption{Participant's evaluation to explanation 5 for run 4}\label{fig:explanation1}
\end{figure*}

\begin{figure*}[h!]%
\centering
\includegraphics[width=0.97\textwidth]{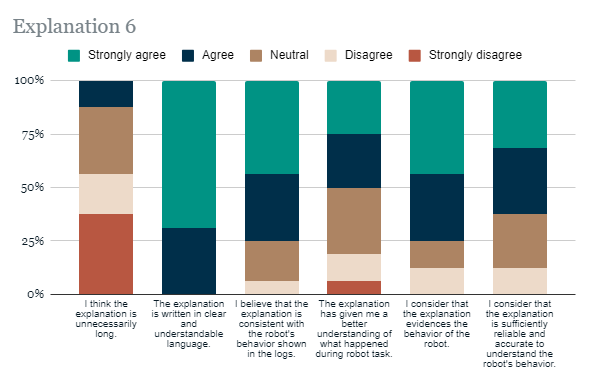}
\caption{Participant's evaluation to explanation 6 for run 4}\label{fig:explanation1}
\end{figure*}

\begin{figure*}[h!]%
\centering
\includegraphics[width=0.97\textwidth]{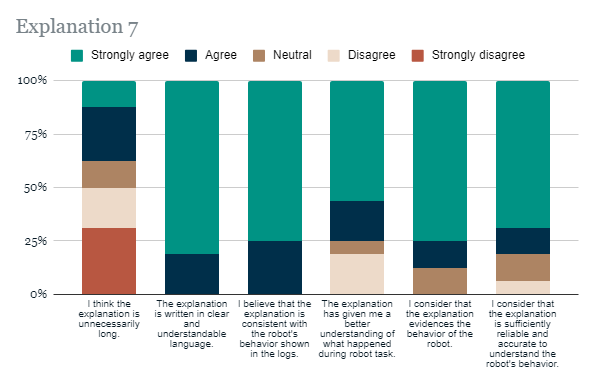}
\caption{Participant's evaluation to explanation 7 for run 4}\label{fig:explanation1}
\end{figure*}

\begin{figure*}[h!]%
\centering
\includegraphics[width=0.97\textwidth]{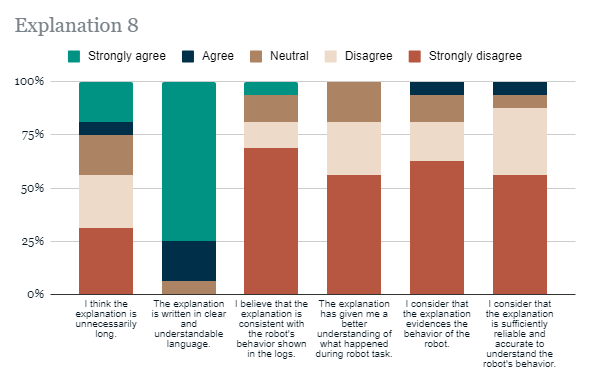}
\caption{Participant's evaluation to explanation 8 for run 4}\label{fig:explanation1}
\end{figure*}

\end{appendices}

\clearpage
\twocolumn
\bibliography{sn-article}

\end{document}